\documentclass[letterpaper, 10 pt, journal, twoside]{IEEEtran}
\usepackage{amsmath,amsfonts}
\usepackage{algorithmic}
\usepackage{array}
\usepackage{textcomp}
\usepackage{stfloats}
\usepackage{url}
\usepackage{cite}
\usepackage{verbatim}
\usepackage{siunitx}
\usepackage{graphicx}
\usepackage{microtype}
\usepackage{cite}
\usepackage{bm}
\usepackage{amsmath,amssymb,amscd,amsfonts,amsthm}
\usepackage{mathtools,leftindex,tensor}
\usepackage[version=4]{mhchem} 
\usepackage{booktabs}
\usepackage{subcaption} 
\usepackage{caption}
\usepackage{soul}

\usepackage[hidelinks]{hyperref}
\usepackage[nameinlink]{cleveref}
\Crefname{figure}{Fig.}{Figs.}
\crefname{section}{Sec.}{Secs.}      
\Crefname{section}{Sec.}{Secs.}      

\usepackage{colortbl}
\usepackage[dvipsnames]{xcolor}
\definecolor{Gray}{gray}{0.85}
\definecolor{LightCyan}{rgb}{0.88,1,1}

\def\BibTeX{{\rm B\kern-.05em{\sc i\kern-.025em b}\kern-.08em
    T\kern-.1667em\lower.7ex\hbox{E}\kern-.125emX}}
\usepackage{balance}
\begin{document}
\title{Multi-Sensor Fusion for Quadruped Robot State Estimation using Invariant Filtering and Smoothing}
\author{Ylenia Nisticò$^{1,*}$, Hajun Kim$^{2,*}$, João Carlos Virgolino Soares$^{1}$, Geoff Fink$^{1,3}$, Hae-Won Park$^{2}$, Claudio Semini$^{1}$
\thanks{Manuscript received: December, 31, 2024; Revised February, 18, 2025; Accepted April, 14, 2025.} 
\thanks{This paper was recommended for publication by Editor Abderrahmane Kheddar upon evaluation of the Associate Editor and Reviewers' comments.

This work was supported by the ASI Pegasus Project and the Technology Innovation Program (00427719, Dexterous and Agile Humanoid Robots for Industrial Applications) funded by the Ministry of Trade Industry \& Energy (MOTIE, Korea).} 

\thanks{$^1$ Ylenia Nisticò, João Carlos Virgolino Soares, and Claudio Semini are with the Dynamic Legged Systems (DLS), Istituto Italiano di Tecnologia (IIT), Genoa, Italy. \tt\footnotesize ylenia.nistico@iit.it, joao.virgolino@iit.it, claudio.semini@iit.it}
\thanks{$^2$ Hajun Kim and Hae-Won Park are with the  Dynamic Robot Control \& Design (DRCD) Laboratory, Korean Advanced Institute of Science and Technology (KAIST), Daejeon, South Korea.
 \tt\footnotesize hajun0219@kaist.ac.kr, haewonpark@kaist.ac.kr 
 }
\thanks{$^3$ Geoff Fink is with the Dynamic Legged Systems (DLS), Istituto Italiano di Tecnologia (IIT), Genoa, Italy, and also with the Department of Engineering, Thompson Rivers University, Kamloops, BC,
Canada. \tt\footnotesize gefink@tru.ca}
\thanks{$^*$Equal contribution. Corresponding author:  \tt\footnotesize ylenia.nistico@iit.it}
\thanks{Digital Object Identifier (DOI): 10.1109/LRA.2025.3564711}
}

\markboth{IEEE Robotics and Automation Letters. Preprint Version. Accepted April, 2025}
{Nisticò \MakeLowercase{\textit{et al.}}:  Multi-Sensor Fusion for Quadruped Robot State Estimation using Invariant Filtering and Smoothing}

\maketitle

\begin{abstract}
This letter introduces two multi-sensor state estimation frameworks for quadruped robots, built on the Invariant Extended Kalman Filter (InEKF) and Invariant Smoother (IS). The proposed methods, named E-InEKF and E-IS, fuse kinematics, IMU, LiDAR, and GPS data to mitigate position drift, particularly along the z-axis, a common issue in proprioceptive-based approaches. We derived observation models that satisfy group-affine properties to integrate LiDAR odometry and GPS into InEKF and IS. LiDAR odometry is incorporated using Iterative Closest Point (ICP) registration on a parallel thread, preserving the computational efficiency of proprioceptive-based state estimation. We evaluate E-InEKF and E-IS with and without exteroceptive sensors, benchmarking them against LiDAR-based odometry methods in indoor and outdoor experiments using the KAIST HOUND2 robot. Our methods achieve lower Relative Position Errors (RPE) and significantly reduce Absolute Trajectory Error (ATE), with improvements of up to 28\% indoors and 40\% outdoors compared to LIO-SAM and FAST-LIO2. Additionally, we compare E-InEKF and E-IS in terms of computational efficiency and accuracy.
\looseness=-1
\end{abstract}


\section{Introduction}
\IEEEPARstart{S}{tate estimation} is crucial for quadruped robots navigating complex environments, as it determines pose, velocity, and key parameters from sensor inputs. Accurate estimation fuses high-frequency proprioceptive inputs (e.g., IMUs, encoders), which suffer from drift, with lower-frequency exteroceptive inputs (e.g., cameras, LiDARs), which may struggle in challenging conditions (e.g., foggy areas, long corridors, etc.).
Sensor fusion techniques address these limitations by combining both sensor types. 
For instance, LIO-SAM~\cite{shan2020lio} employs a factor graph for LiDAR-inertial odometry (LIO), while FAST-LIO~\cite{fastlio2021xu} uses an iterated Extended Kalman Filter (EKF) to fuse LiDAR and IMU data, with FAST-LIO2~\cite{xu2022fast} further improving accuracy by eliminating the need for feature extraction.
\looseness=-1

Although effective, these methods do not explicitly consider contact interactions, which can play a significant role in legged robot locomotion.
Specifically for legged robots, state estimation incorporating leg kinematics has been studied in previous works, such as~\cite{bloesch2013state, bloesch2013stateslippery}, which introduced EKF-based frameworks combining IMU and kinematic measurements under the stable contact assumption. To handle violations of this assumption, a slip rejection method that adjusts covariances during slipping was proposed in~\cite{kim2021slip}.

Among more recent research works, Barrau and Bonnabel proposed in~\cite{barrau2017inekf} the use of the group-affine property for state estimation, which enables a system to be formulated as a linear system by using the Lie algebra error of the state variables defined in a Lie group. They also introduced the invariant EKF (InEKF), which leverages group-affine properties and shows better convergence than the standard EKF when initial errors are present. Hartley et al.~\cite{hartley2020contact} introduced a Proprioceptive-only InEKF (P-InEKF) for legged robots, demonstrating that fusing IMU data with the robot's kinematics results in a state estimation process that is approximately group-affine. Further studies in P-InEKF have explored the incorporation of a learning-based contact estimator~\cite{lin2021legged} or the use of robust scale-variant cost functions to handle challenging terrains~\cite{santana2024proprioceptive}. Moreover, while the P-InEKF processes only the two most recent states, the study in~\cite{chauchat2018smoothing} introduced the Invariant Smoother (IS), which leverages the group-affine property within a smoothing framework that preserves the state history. Yoon et al.~\cite{is_legged} introduced a Proprioceptive-only Invariant Smoother (P-IS) for legged robots, demonstrating robust performance during dynamic contact events while accounting for contact loop costs.

Despite these advancements, proprioceptive-only methods remain prone to drift over time due to an unobservable absolute position and yaw angle~\cite{bloesch2013state}. 
Several state estimation frameworks for legged robots integrate proprioceptive and exteroceptive sensors to address this issue. For instance, Pronto~\cite{pronto2020camurri} uses an EKF to fuse IMU, kinematics, stereo vision, and LiDAR for pose corrections. MUSE~\cite{nistico2025muserealtimemultisensorstate} fuses a nonlinear attitude observer with a Kalman filter for precise state estimation, while STEP~\cite{kim2022step} integrates pre-computed foot velocities and stereo camera data, removing the need for contact detection and stable contact assumptions.
VILENS~\cite{wisth2022vilens} and Cerberus~\cite{cerberus2023yang} leverage factor graphs to robustly fuse multiple sensors, with Cerberus also using visual data for kinematic parameter estimation.
Leg-KILO~\cite{ou24legkilo} tightly integrates IMU, kinematics, LiDAR odometry, and loop closure via graph optimization for enhanced accuracy. 
\looseness=-1

However, the studies in~\cite{pronto2020camurri,nistico2025muserealtimemultisensorstate,kim2022step,wisth2022vilens,cerberus2023yang,ou24legkilo} do not consider state estimation with the group-affine property, despite its advantages in improving estimation accuracy and convergence stability~\cite{barrau2017inekf,chauchat2018smoothing}.
Achieving these benefits requires that both the propagation and observation models for sensor measurements are formulated to satisfy the group-affine conditions~\cite{barrau2017inekf,chauchat2018smoothing}. 

Meanwhile, some studies have explored learning-based methods for state estimation~\cite{zhang2021imu, revach2022kalmannet}. For legged robots,~\cite{youm2024legged} uses a neural measurement model with proprioception, while other works use transformers for the estimation~\cite{schperberg2024optistate, yu2024state}.
However, these techniques require large datasets~\cite{kim2024diter++}, have high computational costs, and may struggle with real-time performance. 
\looseness=-1

Some studies have incorporated exteroceptive sensors into state estimation using the group-affine property. The study in~\cite{teng2021cassie} fuses inertial, kinematics, and linear velocity measurements from a tracking camera, capitalizing on the underlying group-affine structure. Similarly, Gao et al.~\cite{gao2022invariant} presented an InEKF for dynamic rigid surfaces, combining RGB-D cameras and contact orientation measurements using ArUco markers. 
\looseness=-1

To the best of our knowledge, neither the InEKF nor the IS frameworks for legged robots have integrated LiDAR or GPS. In this work, we aim to design an observation model for these frameworks that integrates LiDAR and GPS, in addition to leg kinematics, while preserving the group-affine structure. The main contributions are as follows:
\looseness=-1
\begin{itemize}
    \item 
    We extended InEKF~\cite{hartley2020contact} and IS~\cite{is_legged} by fusing kinematics and IMU, with LiDAR odometry and GPS measurements to mitigate the position drift inherent in proprioceptive-only methods, naming our approaches as \textbf{E-InEKF} and \textbf{E-IS}. To the best of our knowledge, this is the first work to incorporate LiDAR odometry and GPS into InEKF and IS for quadruped robots.
    \item 
    To integrate LiDAR odometry and GPS into the InEKF and the IS, we derive an observation model that satisfies group-affine properties for both sensor types. Additionally, to manage LiDAR's low frequency of approximately 10~Hz, we calculate LiDAR odometry in a parallel thread using the ICP registration of~\cite{vizzo2023ral}, allowing the estimator thread to maintain a fast computation time.
    \item Our algorithms were verified on the KAIST Hound2 quadruped robot~\cite{hound22shin}
    in indoor and outdoor environments, benchmarking the results against LIO-SAM~\cite{shan2020lio} and FAST-LIO2~\cite{xu2022fast}, two LIO systems. Additionally, ablation studies were conducted to evaluate the performance without LiDAR or GPS, and to compare the accuracy and computation trade-offs between E-IS and E-InEKF.  
\end{itemize}

The paper is organized as follows: \Cref{sec:preliminary} outlines the preliminaries for group affine property, P-InEKF, and P-IS. \Cref{sec:method} details the implementation of the observation models in E-InEKF and E-IS. \Cref{sec:experiments} describes  the experiments, while \Cref{sec:discussion} discusses the results. \Cref{sec:conclusion} concludes the paper.

\section{Preliminaries}
\label{sec:preliminary}
This section provides the theoretical background based on~\cite{sola2018micro,barrau2017inekf,chauchat2018smoothing}, and an overview of the P-InEKF~\cite{hartley2020contact} and P-IS~\cite{is_legged}. 

\subsection{Matrix Lie Groups and group-affine property}

In robots' state estimation, the state can be represented on a manifold rather than a vector space. For instance, the orientation is represented by the special orthogonal group $\mathbf{SO}\mathrm{(3)} = \{\mathbf{R} \in \mathbb{R}^{3 \times 3} \mid \det(\mathbf{R}) = 1, \mathbf{R}^\top \mathbf{R} = \mathbb{I}_3\}$, where $\mathbb{I}_\mathrm{nd} \in \mathbb{R}^{\mathrm{nd}\times \mathrm{nd}}$ is the identity matrix. 
When the state includes additional vector components, such as the position or velocity, the elements of  $\mathbf{SE_{k}}\mathrm{(3)}$ can be adopted, which is represented by the $(3+k)\times(3+k)$ matrix and defined as follows~\cite{hartley2020contact}:
\begin{equation}
    \mathbf{X} \triangleq \begin{bmatrix}
        \mathbf{R} & \prescript{1}{}{p} &  ... & \prescript{k}{}{p} \\ 
        \bm{0}_{k,3} &  &    \mathbb{I}_{k}     & &         
        \end{bmatrix}\in\mathbf{SE_{k}}\mathrm{(3)} ,
    \label{eq:state_matrix_se3}
\end{equation}
where $\mathbf{R} \in \mathbf{SO}\mathrm{(3)}$ is the rotation matrix, and $\prescript{i}{}{p} \in \mathbb{R}^3$, for $i=1,2,\ldots,k$, is a vector. The element $\mathbf{X} \in \mathbf{SE}_{k}\mathrm{(3)}$ can be mapped to $\xi\triangleq[\phi^{\top},{^{1}}\xi^{\top},\cdots,{^{k}}\xi^{\top}]^{\top}\in\mathbb{R}^{3+3k}$, which lies in an Euclidean vector space, via the logarithmic map represented by $\mathbf{X} \in  \mathbf{SE}_{k}\mathrm{(3)} \rightarrow \mathrm{Log}(\mathbf{X})  \in \mathbb{R}^{3+3k}$ and back via the exponential map described by $ \mathbf{\xi}   \in \mathbb{R}^{3+3k}  \rightarrow  \mathrm{Exp}(\mathbf{\xi})  \in \mathbf{SE}_{k}\mathrm{(3)}$.
The exponential map in $\mathbf{SE}_{k}\mathrm{(3)}$ is defined as follows:
\begin{equation}  
\mathrm{Exp}(\mathbf{\xi}) = \mathrm{exp}(\mathbf{\xi}^{\wedge})= \begin{bmatrix}
        \mathrm{exp}(\phi^{\wedge}) & \mathbf{J}_{l}(\mathrm{\phi})\prescript{1}{}{\xi} &  \cdots & \mathbf{J}_{l}(\mathrm{\phi})\prescript{k}{}{\xi} \\ 
        \bm{0}_{k,3} &   & \mathbb{I}_{k}        & 
    \end{bmatrix},
    \label{eq:exponential_map_se3_definition}
\end{equation}
where $\mathbf{J}_{l}(\cdot)$ is the left Jacobian of the $\mathbf{SO}\mathrm{(3)}$ manifold, and the $\textit{hat}$ operation $(\cdot)^{\wedge}$ on 
$\phi$ and $\xi$ , which is the inverse mapping of the $\textit{vee}$ operation $(\cdot)^{\vee}$, is respectively defined as
\begin{equation}
    \mathbf{\phi}^{\wedge} =\begin{bmatrix}
        0 & -\phi_\mathrm{z} & \phi_\mathrm{y} \\ 
    \phi_\mathrm{z} & 0 & -\phi_\mathrm{x} \\
    -\phi_\mathrm{y} & \phi_\mathrm{x} & 0\\
    \end{bmatrix},
    \label{eq:hat_operation}
    \mathbf{\xi}^{\wedge}=\begin{bmatrix}
        \phi^{\wedge}& {^{1}\xi} &\cdots &{^{k}\xi} \\
        \mathbf{0}_{k,3}& \mathbf{} &\mathbf{0}_{k,k} &\mathbf{}
    \end{bmatrix}
\end{equation}
The adjoint matrix, which plays a key role in transforming perturbations in the tangent space, is defined as: 
\begin{equation}  
\text{Ad}_\mathbf{X}=
\begin{bmatrix} 
    \mathbf{R} & \mathbf{0}_{3\times3} &  \cdots & \mathbf{0}_{3\times3} \\ 
    \prescript{1}{}{p}^{\wedge}\mathbf{R} & \mathbf{R} & \cdots & \mathbf{0}_{3\times3}\\
    \vdots & \vdots & \ddots & \vdots \\
        \prescript{k}{}{p}^{\wedge}\mathbf{R} & \mathbf{0}_{3\times3} &  \cdots     & \mathbf{R}  
\end{bmatrix}.
    \label{eq:adjoint_matrix_definition}
\end{equation}
Also, we define the $\odot$ operation, introduced in~\cite{is_legged}, as follows:
\begin{equation}
    \xi^{\odot}\triangleq\begin{bmatrix}
        -\phi^{\wedge}& {^{1}}\xi\cdot{\mathbb{I}_{3}}& \cdots& {^{k}}\xi\cdot{\mathbb{I}_{3}}\\
        \mathbf{0}_{k,3}& \mathbf{0}_{k,3}& \cdots& \mathbf{0}_{k,3}
    \end{bmatrix}\in\mathbb{R}^{(3+k)\times(3+3k)}.
    \label{eq:odot}
\end{equation}

The group-affine property provides a significant advantage by ensuring that the observer's error dynamics do not depend on the current state when the propagation and observation models, defined on the Lie Group, are group-affine~\cite{barrau2017inekf}.

For an arbitrary state $\mathbf{X}_\mathrm{t}$ at time $\mathrm{t}$ and its corresponding estimate $\bar{\mathbf{X}}_\mathrm{t}$ in the Lie Group $\mathbb{G}$, the right-invariant error is defined as $\eta^{\mathrm{r}}_\mathrm{t}\triangleq\bar{\mathbf{X}}^{-1}_\mathrm{t}\mathbf{X}_\mathrm{t}$, and the corresponding right log-invariant error is $\xi^{\mathrm{r}}_\mathrm{t}\triangleq\mathrm{Log}(\eta^{r}_\mathrm{t}).$ A system is group-affine if its propagation model $f(\cdot)$ satisfies the following conditions:
$\frac{\mathrm{d}}{\mathrm{dt}} \mathbf{X}_\mathrm{t} = f(\mathbf{X}_\mathrm{t})$ and $f(\mathbf{XY}) = \mathbf{X}f(\mathbf{Y}) + f(\mathbf{X})\mathbf{Y} - \mathbf{X}f(\mathbb{I}_{\text{dim}(\mathbf{Y})})\mathbf{Y}$ $\text{for} \quad \forall \mathrm{t} \geq 0 \quad \text{and} \quad \forall \mathbf{X}, \mathbf{Y} \in \mathbb{G}$.  

When the trajectories $\mathbf{X}_{0:\mathrm{t}}$ and $\bar{\mathbf{X}}_{0:\mathrm{t}}$ are governed by the same group-affine propagation, the right invariant error's propagation dynamics is $\frac{d}{dt}\eta^{r}_{\mathrm{t}}=g(\eta^{r}_{\mathrm{t}})$, where a function $g(\cdot)$ is represented by $g(\eta^{r}_{\mathrm{t}})=f(\eta^{r}_{\mathrm{t}})-\eta^{r}_{\mathrm{t}}f(\mathbb{I}_{\text{dim}(\mathbf{Y})})$, as shown in~\cite{barrau2017inekf}. For small errors, it is also proven in~\cite{barrau2017inekf} that the propagation of the log-invariant error $\xi^{\mathrm{r}}_\mathrm{t}$ is a linear differential equation, with a constant matrix $\mathbf{G}$: 
\looseness=-1
\begin{equation}
g(\eta^{\mathrm{r}}_\mathrm{t}) = (\mathbf{G}\xi^{\mathrm{r}}_\mathrm{t})^\wedge + \mathcal{O}(\|\xi^{\mathrm{r}}_\mathrm{t}\|^2), \quad \frac{\mathrm{d}}{\mathrm{dt}}\xi^{\mathrm{r}}_\mathrm{t} = \mathbf{G}\xi^{\mathrm{r}}_\mathrm{t},
\label{eq:prop_group_affine_property_G}
\end{equation}
where $\mathcal{O}({\cdot})$ indicates higher order terms that are negligible. Due to the constant matrix $\mathbf{G}$, the evolution of the error $\xi^{r}_{t}$ is autonomous and independent of the two trajectories $\mathbf{X}_{0:\mathrm{t}}$ and $\bar{\mathbf{X}}_{0:\mathrm{t}}$, which is referred to as the \textit{log-linear property} of the error~\cite{barrau2017inekf}.

Meanwhile, the group-affine property for the observation model is satisfied under the condition:
\begin{equation}
\mathbf{y}_\mathrm{t} = \mathbf{X}_\mathrm{t}^{-1} \mathbf{s} + \mathbf{w}^{\text{obs}}_\mathrm{t} \quad, 
\label{eq:obs_group_affine_property}
\end{equation}
where $\mathbf{y}_\mathrm{t}$ is the observation at time $\mathrm{t}$, $\mathbf{s}$ is a constant vector, and $\mathbf{w}^{\mathrm{obs}}_\mathrm{t}$ is Gaussian noise vector of the observation. 

\subsection{State Definition}
Inspired by \cite{hartley2020contact,is_legged}, we define the state \textbf{$\mathbf{X}_\mathrm{t}$} $\in \mathbf{SE}_\mathrm{{N+2}}(3)$ for $N$ contact points, while the IMU bias vector \textbf{$\mathbf{x}_\mathrm{t}$} $\in \mathbb{R}^6$ is defined in the vector space as follows: 

\begin{equation}
    \mathbf{X}_{\mathrm{t}} \triangleq \begin{bmatrix}
        \mathbf{R}_\mathrm{t} &  \mathbf{v}_\mathrm{t} &\mathbf{p}_\mathrm{t} \ \ \mathbf{d}_\mathrm{1,t} \ \cdots \ \mathbf{d}_\mathrm{N,t} \\ 
        0_{N+2,3} & &\mathbb{I}_{(N+2)}   
    \end{bmatrix} 
    \quad \text{and} \quad
    \mathbf{x}_{\mathrm{t}} \triangleq \begin{bmatrix}
        \mathbf{b}^{\omega}_\mathrm{t} \\ 
        \mathbf{b}^\mathrm{a}_\mathrm{t}
    \end{bmatrix},
    \label{eq:state_matrix}
\end{equation}
where $\mathbf{R}_\mathrm{t}$ is the base orientation, $\mathbf{v}_\mathrm{t}$ is the linear 
velocity, $\mathbf{p}_\mathrm{t}$ is the robot position, and $\mathbf{d}_\mathrm{i,t}$ is $i$-th contact position, while $\mathbf{b}^\mathrm{\omega}_\mathrm{t}$, and $\mathbf{{b}}^\mathrm{a}_\mathrm{t}$ are the IMU biases for gyroscope and accelerometer.
\looseness = -1
 
For the sake of readability, since the measurement models for each contact point $\mathbf{d}_\mathrm{i,t}$ are identical, we will derive all further equations assuming only a single contact point $\mathbf{d}_\mathrm{t}$.

We also consider the measurements
as $\mathbf{Z}_\mathrm{t}\triangleq\left[\boldsymbol{\tilde{\omega}}_\mathrm{t}^\top, \mathbf{\tilde{a}}_\mathrm{t}^{\top}, 
\mathbf{\tilde{q}^\top}_\mathrm{t}, \mathbf{\dot{\tilde{q}}^{\top}_\mathrm{t}},
\mathbf{p}_\mathrm{cl,t}^{\top},
\mathbf{p}_\mathrm{gps,t}^\top\right]$,
\noindent
 where $\boldsymbol{\tilde{\omega}}_\mathrm{t} \in \mathbb{R}^3$, 
$\mathbf{\tilde{a}}_\mathrm{t} \in \mathbb{R}^3$, $\mathbf{\tilde{q}}_\mathrm{t} 
\in \mathbb{R}^{3M}$, $\mathbf{\dot{\tilde{q}}}_\mathrm{t} \in \mathbb{R}^{3M}$, $\boldsymbol{\mathbf{p}}_\mathrm{cl,t}$, and $\boldsymbol{\mathbf{p}}_\mathrm{gps,t}$  are the angular velocity, the linear acceleration, joint positions, joint velocities for the $M$-legged robot, where $M$ is the number of legs, LiDAR point clouds, and position data from GPS, respectively.

\subsection{Invariant Extended Kalman Filter For Legged Robots}
\label{subsec:propagation}
In the InEKF~\cite{hartley2020contact}, both the propagation and observation models are formulated under the group-affine property~\cite{hartley2020contact} to ensure that the right-invariant error is independent of the current state. The propagation model integrates IMU measurements to predict the state, whereas the observation model refines the estimates under the stable contact assumption, which implies zero foot velocity during ground contact.

The system dynamics for the propagation that includes the corruption from noise and bias can be described as follows~\cite{hartley2020contact}:
\looseness=-1
\begin{align}
        \frac{\mathrm{d}}{\mathrm{dt}}\mathbf{R}_{\mathrm{t}} &= \mathbf{R}_{\mathrm{t}}(\mathbf{\tilde{\omega}}_{\mathrm{t}}-\mathbf{b}^\mathbf{\omega}_{\mathrm{t}} - \mathbf{w}^{\mathbf{\omega}}_\mathrm{t})^{\wedge}, \quad  
        \frac{\mathrm{d}}{\mathrm{dt}}\mathbf{p}_{\mathrm{t}} = \mathbf{v}_{\mathrm{t}}
        \label{eq:dyn1}
        \\
        \frac{\mathrm{d}}{\mathrm{dt}}\mathbf{v}_{\mathrm{t}} &= \mathbf{R}_{\mathrm{t}}(\mathbf{\tilde{a}}_{\mathrm{t}} - \mathbf{b}^{\mathbf{a}}_{\mathrm{t}} - \mathbf{w}^\mathbf{a}_\mathrm{t}) + \mathbf{g},\quad \frac{\mathrm{d}}{\mathrm{dt}}\mathbf{d}_{\mathrm{t}}  = \mathbf{R}_{\mathrm{t}} \mathbf{w}^\mathbf{d}_{\mathrm{t}}\label{eq:dyn2} \\
        \frac{\mathrm{d}}{\mathrm{dt}}\mathbf{b}^{\mathbf{\omega}}_{\mathrm{t}} &= \mathbf{w}^{\mathbf{b}^\mathbf{\omega}}_\mathrm{t}, \quad \frac{\mathrm{d}}{\mathrm{dt}}\mathbf{b}^{\mathbf{a}}_{\mathrm{t}} = \mathbf{w}^{\mathbf{b}^\mathbf{a}}_\mathrm{t} \label{eq:dyn3}  
\end{align} 
\noindent
where $\mathbf{g}$ is the gravity vector, $\mathbf{w^\omega_{\mathrm{t}}}$, $\mathbf{w^a_{\mathrm{t}}}$, $\mathbf{w^d_{\mathrm{t}}}$, $\mathbf{w^{b^\omega}_{\mathrm{t}}}$, and $\mathbf{w^{b^a}_{\mathrm{t}}}$ are the zero-mean Gaussian noise terms of each process.
The propagation model can be described as the noise-corrupted version in the discrete domain~\cite{hartley2020contact} of the log-linear property~\eqref{eq:prop_group_affine_property_G}:
\begin{equation}
\begin{bmatrix}\xi^{\mathrm{r}}_\mathrm{t+1}\\
\zeta^{\mathrm{r}}_\mathrm{t+1}
\end{bmatrix}
= (\mathbb{I}_{18}+\mathbf{A}_{\mathrm{t}}\Delta{t}) 
\begin{bmatrix}
\xi^{\mathrm{r}}_\mathrm{t}\\
\zeta^{\mathrm{r}}_\mathrm{t}
\end{bmatrix}
+\mathbf{B}_{\mathrm{t}}\bar{\mathbf{w}}_\mathrm{t},
\mathbf{B}_{\mathrm{t}}=\Delta{t}
\begin{bmatrix}
\text{Ad}_{\bar{\mathbf{X}}_\mathrm{t}} \quad \mathbf{0}_{12,6} \\
\mathbf{0}_{6,12} \quad  \mathbb{I}_{6}
\end{bmatrix}
    \label{eq:log_linear_prop}
\end{equation}
\begin{equation}
\mathbf{A}_{\mathrm{t}}=\begin{bmatrix}\mathbf{0} & \mathbf{0}_{3,3} & \mathbf{0}_{3,3} & \mathbf{0}_{3,3} & -\overline{\mathbf{R}}_\mathrm{t} & \mathbf{0}_{3,3} \\ (\mathbf{g})^{\wedge} & \mathbf{0}_{3,3} & \mathbf{0}_{3,3} & \mathbf{0}_{3,3} & -\left(\overline{\mathbf{v}}_\mathrm{t}\right)^{\wedge} \overline{\mathbf{R}}_\mathrm{t} & -\overline{\mathbf{R}}_\mathrm{t} \\ \mathbf{0}_{3,3} & \mathbb{I}_{3} & \mathbf{0}_{3,3} & \mathbf{0}_{3,3} & -\left(\overline{\mathbf{p}}_\mathrm{t}\right)^{\wedge} \overline{\mathbf{R}}_\mathrm{t} & \mathbf{0}_{3,3} \\ \mathbf{0}_{3,3} & \mathbf{0}_{3,3} & \mathbf{0}_{3,3} & \mathbf{0}_{3,3} & -\left(\overline{\mathbf{d}}_\mathrm{t}\right)^{\wedge} \overline{\mathbf{R}}_\mathrm{t} & \mathbf{0}_{3,3} \\ \mathbf{0}_{6,3} & \mathbf{0}_{6,3} & \mathbf{0}_{6,3} & \mathbf{0}_{6,3} & \mathbf{0}_{6,3} & \mathbf{0}_{6,3} 
\end{bmatrix},
    \label{eq:A_matrix_IS}
\end{equation}
where $\Delta{t}$ is the time step of the state estimation, $\bar{\mathbf{w}}_\mathrm{t} = [
(\mathbf{w}^{\omega}_\mathrm{t})^{\top}$,$ (\mathbf{w}^{\mathbf{a}}_\mathrm{t})^{\top}$,$ (\mathbf{w}^{\mathbf{a}}_\mathrm{t}\Delta{t})^{\top}$,$ (\mathbf{w}^{\mathbf{d}}_\mathrm{t})^{\top}$,$(\mathbf{w}^{b^{\omega}}_{t})^{\top}$,$(\mathbf{w}^{b^{a}}_{t})^{\top}]^{\top}$ is the noise vector, $\mathbf{e}_{\mathrm{t}}\triangleq\begin{bmatrix}
    (\xi^{r}_{\mathrm{t}})^{\top},(\zeta^{r}_{\mathrm{t}})^{\top}
\end{bmatrix}^{\top}$, and the bias error can be defined as $\zeta^{r}_{\mathrm{t}}=\mathbf{x_{t}}-\bar{\mathbf{x_{t}}}$.

For the observation model, the forward kinematic measurement, which captures the relative position of the contact point with respect to the body, is formulated to satisfy the group-affine property of~\eqref{eq:obs_group_affine_property}. 
The forward kinematics $\mathrm{fk}(\mathbf{\tilde{q}}_\mathrm{t})$ can be expressed as: $\mathrm{fk}(\mathbf{\tilde{q}}_\mathrm{t}) = \mathbf{R}^\top_\mathrm{t}(\mathbf{d}_\mathrm{t}
-\mathbf{p}_\mathrm{t})+\mathbf{J}_\mathrm{p}(\mathbf{\tilde{q}}_\mathrm{t})\mathbf{w}^{q}_\mathrm{t}$, where $\mathbf{J}_\mathrm{p}$ denotes the analytical Jacobian of the forward kinematics, and $\mathbf{w}^{q}_\mathrm{t}$ is the observation noise in forward kinematics.
As in~\eqref{eq:obs_group_affine_property}, the right-invariant form of this measurement can be described as follows, allowing the innovation to remain only dependent on the invariant error~\cite{hartley2020contact}:
\begin{equation}
    \mathbf{Y}^{\mathrm{kin}}_{\mathrm{t}} = \mathbf{X}^{-1}_{\mathrm{t}} \mathbf{b}^{\mathrm{kin}}_{\mathrm{t}} + \mathbf{V}^{\mathrm{kin}}_{\mathrm{t}},
    \label{eq:kine_meas}
\end{equation}
where $\mathbf{Y}^{\mathrm{kin}}_{\mathrm{t}} = [\mathrm{fk}(\mathbf{\tilde{q}}_\mathrm{t}) ~ 0 ~ 1 ~ -1]^\top$ is the vector of kinematic observation, $\mathbf{X}^{-1}_{\mathrm{t}}$ is the inverse of the state matrix~$\mathbf{X}_{\mathrm{t}}$,  
${\mathbf{b}^{\mathrm{kin}}_{\mathrm{t}}} = [\mathbf{0}_{3,1} ~ 0 ~ 1 ~ -1]^\top $ is a constant vector, while ${\mathbf{V}^{\mathrm{kin}}_{\mathrm{t}}} = [\mathbf{J_p}(\mathbf{\tilde{q}_t}) \mathbf{w_t^q} ~ 0 ~ 0 ~ 0]^\top$ 
is the Gaussian noise vector of the observation model.

Given the observation model of~\eqref{eq:obs_group_affine_property}, which satisfies the group-affine property, the state and covariance updates of the observation model can be expressed as follows~\cite{hartley2020contact}: $\bar{\mathbf{X}}^{+}_{t}=\mathrm{Exp}(\mathbf{K}_{\mathrm{t}}\Pi\bar{\mathbf{X}}^{-}_{\mathrm{t}}\mathbf{Y}_{\mathrm{t}})\bar{\mathbf{X}}^{-}_{\mathrm{t}}$, $\mathbf{P}^{+}_{t}=(\mathbb{I}-\mathbf{K}_{\mathrm{t}}\mathbf{H}_{\mathrm{t}})\mathbf{P}_{\mathrm{t}}(\mathbb{I}-\mathbf{K}_{\mathrm{t}}\mathbf{H}_{\mathrm{t}})^{\top}+\mathbf{K}_{\mathrm{t}}\bar{\mathbf{N}}_{\mathrm{t}}\mathbf{K}^{\top}_{\mathrm{t}}$
, where $\bar{\mathbf{X}}^{-}_{\mathrm{t}}$ is the estimated state from system dynamics, $\mathbf{P}_{t}$ is the system covariance matrix, and $\Pi\triangleq\begin{bmatrix}\mathbb{I} \quad\mathbf{0}_{3,3}\end{bmatrix}$ is the auxiliary selection matrix. The Kalman gain matrix $\mathbf{K}_{\mathrm{t}}$ is computed as $\mathbf{K}_{\mathrm{t}} =\mathbf{P}_{\mathrm{t}}\mathbf{H}_{\mathrm{t}}^{\top}(\mathbf{H}_{\mathrm{t}}\mathbf{P}_{\mathrm{t}}\mathbf{H}^{\top}_{\mathrm{t}}+\bar{\mathbf{N}}_{\mathrm{t}})^{-1}$. The matrices $\mathbf{H}_{\mathrm{t}}$ and $\bar{\mathbf{N}}_{\mathrm{t}}$ are given by $\mathbf{H}_{\mathrm{t}}=\begin{bmatrix}
\mathbf{0}_{3,3} \quad \mathbf{0}_{3,3} \quad -\mathbb{I} \quad \mathbb{I}   
\end{bmatrix}$ and $\bar{\mathbf{N}}_{\mathrm{t}}=\mathbf{R}^{-}_{\mathrm{t}}\mathbf{J_{p}}(\mathbf{\tilde{q}_\mathrm{t}})\text{Cov}(\mathbf{w}^{\mathbf{q}}_{\mathrm{t}})\mathbf{J_{p}}^{\top}(\mathbf{\tilde{q}_\mathrm{t}})(\mathbf{R}^{-}_{\mathrm{t}})^{\top}.$
\looseness=-1

\subsection{Invariant Smoother For Legged Robots}
Unlike the filtering method of~\cite{hartley2020contact}, which 
handles the last two states 
of the system, the smoothing method aims to recover the estimates using Maximum A Posteriori (MAP) estimation, by incorporating the states $\mathbf{X}_\mathrm{0:n}$ and measurements $\mathbf{Z}_\mathrm{0:n}$ within the specific time window $\mathrm{n}$.
As derived in~\cite{is_legged}, MAP can be formulated as a nonlinear least-squares optimization problem as follows: 
\looseness=-1
\begin{equation}
\begin{aligned}
    &\operatorname*{\mathbf{e}^{*}}_{0:\mathrm{n}} 
    = \operatorname*{argmin}_{\mathbf{e}_{0:\mathrm{n}}} ({\| \mathbf{r}_{\text{pri}} - \mathbf{J}_{\text{pri}}\mathbf{e}_{0:\mathrm{n}} \|^2_{\Sigma_{\text{pri}}}} 
    \\ &+ {\sum_{{{\mathrm{t}}}=0}^{\mathrm{n}-1} \| \mathbf{r}^{{\mathrm{t}}}_{\text{pro}}-\mathbf{J}^{{\mathrm{t}}}_{\text{pro}}\mathbf{e}_{0:n}\|^2_{\Sigma_{\text{pro}}}}  
    + {\sum_{{{\mathrm{t}}}=0}^{\mathrm{n}} \| \mathbf{r}^{{\mathrm{t}}}_{\text{o}}-\mathbf{J}^{{\mathrm{t}}}_{\text{o}}\mathbf{e}_{0:n} \|^2_{\Sigma_{\text{o}}}}),
    \end{aligned}
\label{eq:is_cost}
\end{equation}
where $\mathrm{n}$ is the window size (WS)
, $\mathbf{r}_{\text{pri}}$, $\mathbf{r}_{\text{pro}}$ and $\mathbf{r}_{\text{o}}$ are residuals, $\Sigma_{\text{pri}}$, $\Sigma_{\text{pro}}$ and $\Sigma_{\text{o}}$ are covariance matrices, and $\mathbf{J}_{\text{pri}}$, $\mathbf{J}_{\text{pro}}$ and $\mathbf{J_{\text{o}}}$ are the Jacobians at the operating points $\bar{\mathbf{X}}_{\mathrm{t}}$ and $\bar{\mathbf{x}}_{\mathrm{t}}$ for the distribution of  \textit{prior}, \textit{propagation}, and \textit{observation}, respectively.
\looseness=-1

The prior cost gives an initialization to state estimation, employing marginalization to keep the fixed size of the time window. The detailed derivation is explained in~\cite{is_legged} and not repeated here. 
\looseness=-1
The propagation cost can be derived from the log-linear property, as shown in~\cite{is_legged}, and expressed as follows: 
\begin{equation}
\mathbf{r}^{\mathrm{t}}_{\text{pro}}=\begin{bmatrix}
        \mathrm{Log}(f^{\mathrm{d}}_{M}(\bar{\mathbf{X}}_{{\mathrm{t}}})\bar{\mathbf{X}}^{-1}_{t+1}) \\ f^{\mathrm{d}}_{v}(\bar{\mathbf{x}}_{{\mathrm{t}}})-\bar{\mathbf{x}}_{{\mathrm{t}}+1}
    \end{bmatrix},
    \Sigma_{\text{pro}}=\mathbf{B}_{\mathrm{t}}\mathrm{Cov}(\bar{\mathbf{w}}_{\mathrm{t}})\mathbf{B}^{\top}_{\mathrm{t}},
\end{equation}
\begin{equation}
\mathbf{J}^{{\mathrm{t}}+1}_{\mathrm{pro}}=\mathbb{I}_{18}+\mathbf{A}_{\mathrm{t}}\Delta{t},
    \quad\mathbf{J}^{{\mathrm{t}}}_{\mathrm{pro}}=\mathbb{I}_{18},
\end{equation}
\begin{equation}
    f^{\mathrm{d}}_{M}({\mathbf{X}}_{{\mathrm{t}}})=\begin{bmatrix}
        \mathbf{R}^{\mathrm{d}}_{{\mathrm{t}}} \quad &\mathbf{v}^{\mathrm{d}}_{{\mathrm{t}}} \quad
        &\mathbf{p}^{\mathrm{d}}_{{\mathrm{t}}} \quad
        &\mathbf{d}^{\mathrm{d}}_{{\mathrm{t}}} \\ \mathbf{0}_{3,3} 
        \quad 
        &\mathbf{}
        \quad
        &\mathbb{I}_{3}
        \quad
        &\mathbf{}
    \end{bmatrix}, 
    f^{\mathrm{d}}_{v}(\mathbf{x}_{{\mathrm{t}}})=\begin{bmatrix}
    \mathbf{b}^{\omega}_{{\mathrm{t}}} \\
    \mathbf{b}^{\mathbf{a}}_{{\mathrm{t}}}
    \end{bmatrix},
\end{equation}
where $f^{\mathrm{d}}_{M}$ is the discretized dynamics of~\eqref{eq:dyn1} and~\eqref{eq:dyn2}, $f^{\mathrm{d}}_{v}$ is the discretized dynamics of \eqref{eq:dyn3}, $\mathbf{R}^{\mathrm{d}}_{{\mathrm{t}}}=\mathbf{R}_{{\mathrm{t}}}\mathrm{Exp}((\bar{\omega}_{{\mathrm{t}}}-\mathbf{b}^{\omega}_{{\mathrm{t}}})\Delta{t})$, $\mathbf{v}^{\mathrm{d}}_{{\mathrm{t}}}=\mathbf{v}_{{\mathrm{t}}}+\mathbf{R}_{{\mathrm{t}}}(\bar{\mathbf{a}}_{{\mathrm{t}}}-\mathbf{b}^{\mathbf{a}}_{{\mathrm{t}}})\Delta{t}+\mathbf{g}\Delta{t}$, and $\mathbf{p}^{\mathrm{d}}_{{\mathrm{t}}}=\mathbf{p}_{{\mathrm{t}}}+\mathbf{v}_{{\mathrm{t}}}\Delta{t}+\frac{1}{2}\mathbf{R}_{{\mathrm{t}}}(\bar{\mathbf{a}}_{{\mathrm{t}}}-\mathbf{b}^{\mathbf{a}}_{{\mathrm{t}}})(\Delta{t})^{2}+\frac{1}{2}\mathbf{g}(\Delta{t})^{2}$.

For the observation cost, the right-invariant kinematics observation of~\eqref{eq:kine_meas} can be employed as in~\cite{is_legged}:
$\mathbf{r}^{{\mathrm{t}}}_{\mathrm{o}}= \mathbf{X}_{{\mathrm{t}}}\mathbf{Y}^{\mathrm{kin}}_\mathrm{t} - \mathbf{b}^{\mathrm{kin}}_{{\mathrm{t}}},
    \mathbf{J_{o}}^\mathrm{t} = [(\mathbf{b}^{\mathrm{kin}}_{\mathrm{t}})^\odot \hspace{0.2cm} \mathbf{0}_{6,6}]
,\space\space\mathbf{\Sigma}^{{\mathrm{t}}}_{\mathbf{o}} = \bar{\mathbf{X}}_{{\mathrm{t}}} \mathbf{\Sigma_{\mathrm{kin}}(w^q)_{\mathrm{t}}} \bar{\mathbf{X}}^\top_{{\mathrm{t}}}$,
where the $\odot$ operator is defined in~\eqref{eq:odot}.

When the MAP of~\eqref{eq:is_cost} is solved, the optimized variables updates the operating states, $\bar{\mathbf{X}}_{i}$ and $\bar{\mathbf{x}}_{i}$, at each iteration as: $\mathbf{X}^{*}_{\mathrm{t}}\leftarrow\mathrm{Exp}(\xi^{r*}_{\mathrm{t}})\bar{\mathbf{X}}_{\mathrm{t}}, \mathbf{x}^{*}_{\mathrm{t}}\leftarrow\bar{\mathbf{x}}_{i}+\zeta^{r*}_{i}.$
\looseness=-1

\section{Proposed Observation Model}
\label{sec:method}
The main contribution of this paper is a measurement model that integrates kinematic measurements with exteroceptive sensor data, specifically LiDAR and GPS, while satisfying the group-affine property of~\eqref{eq:obs_group_affine_property}, thereby enabling the incorporation of sensor measurements into InEKF or IS.
\looseness=-1
\subsection{LiDAR Odometry Factor}
For LiDAR odometry, we adopt KISS-ICP~\cite{vizzo2023ral}, an efficient algorithm that estimates the pose by sequentially aligning LiDAR point clouds using a point-to-point ICP method. 
KISS-ICP applies a constant-velocity motion model to deskew scans, uses voxel-based downsampling to reduce computational load, adapts thresholds for data association based on motion, and refines pose estimates through robust optimization handling point-to-point ICP. By solving the ICP problem, we obtain the position of the LiDAR odometry, $\mathbf{p_{lid}}$, in the world frame.

Unlike many other odometry techniques, KISS-ICP operates independently of additional sensors such as IMUs or leg odometry, enhancing adaptability. Furthermore, we employed a loosely coupled
approach, running the LiDAR odometry on a separate thread to preserve the efficiency of proprioceptive state estimation. 
This loosely coupled setup allows KISS-ICP to deliver reliable LiDAR-based position updates, which integrate seamlessly into our system. 

\subsection{Observation Model for LIDAR and GPS}
\subsubsection{{LiDAR measurement model}}
by computing the LiDAR position estimate $\mathbf{p_{lid}}$ from the point-to-point ICP problem~\cite{vizzo2023ral},
we express the LiDAR position estimate $\mathbf{p_{{lid}}}$ with respect to the state $\mathbf{X}_\mathrm{t}$ in the right-invariant form as given in~\eqref{eq:obs_group_affine_property}:
\begin{equation}
    \mathbf{Y_{lid}} = \mathbf{X}^{-1}_\mathrm{t}\mathbf{b_{lid}}+\mathbf{V_{lid}}
    \label{eq:lid_meas}
\end{equation}
where $\mathbf{b_{lid}} = [\mathbf{p_{lid}} ~ 0 ~ 1 ~ 0]^\top$ is the LiDAR observation vector, 
${\mathbf{Y_{lid}}} = [\mathbf{0}_{3,1} ~ 0 ~ 1 ~ 0]^\top$ is a constant vector, while ${\mathbf{V_{lid}}} = [\mathbf{J}_{\mathbf{p_{lid}}} \mathbf{w_t^{lid}} ~ 0 ~ 0 ~ 0]^\top$ 
is the Gaussian noise vector associated with the observation model.
This formulation aligns LiDAR measurements with the group-affine property, facilitating consistent integration within the invariant estimation frameworks.
\looseness=-1

\subsubsection{{GPS measurement model}}
to incorporate GPS position estimates $\mathbf{p_{gps}}$ into the invariant state estimation framework, we express the GPS position $\mathbf{p_{gps}}$ with respect to the state $\mathbf{X}_\mathrm{t}$ as a right-invariant observation model:
\begin{equation}
    \mathbf{Y_{gps}} = \mathbf{X}^{-1}_\mathrm{t} \mathbf{b_{gps}} + \mathbf{V_{gps}}
    \label{eq:gps_meas}
\end{equation}
where $\mathbf{b_{gps}} = [\mathbf{p_{gps}} ~ 0 ~ 1 ~ 0]^\top$ represents the GPS observation vector, 
${\mathbf{Y_{gps}}} = [\mathbf{0}_3,1 ~ 0 ~ 1 ~ 0]^\top$ is a constant vector, while ${\mathbf{V_{gps}}} = [\mathbf{J}_{\mathbf{p_{gps}}} \mathbf{w_t^{gps}} ~ 0 ~ 0 ~ 0]^\top$ 
is the Gaussian noise vector associated with the observation model.

Based on the measurement model corrupted by zero-mean Gaussian noise, we derived the observation models for LiDAR and GPS measurements. The introduced observation models for LiDAR and GPS of \Cref{eq:lid_meas,eq:gps_meas} fulfill the group-affine property required for the invariant state estimation, because our measurement models follow the right-invariant observation form~\cite{barrau2017inekf}, given in~\Cref{eq:obs_group_affine_property}. Thus, our approach preserves the invariant structure, ensuring that the error dynamics of the estimator remain autonomous and independent of the robot's trajectory.

\begin{figure}[t]
    \centering
    \includegraphics[width=0.45\textwidth]{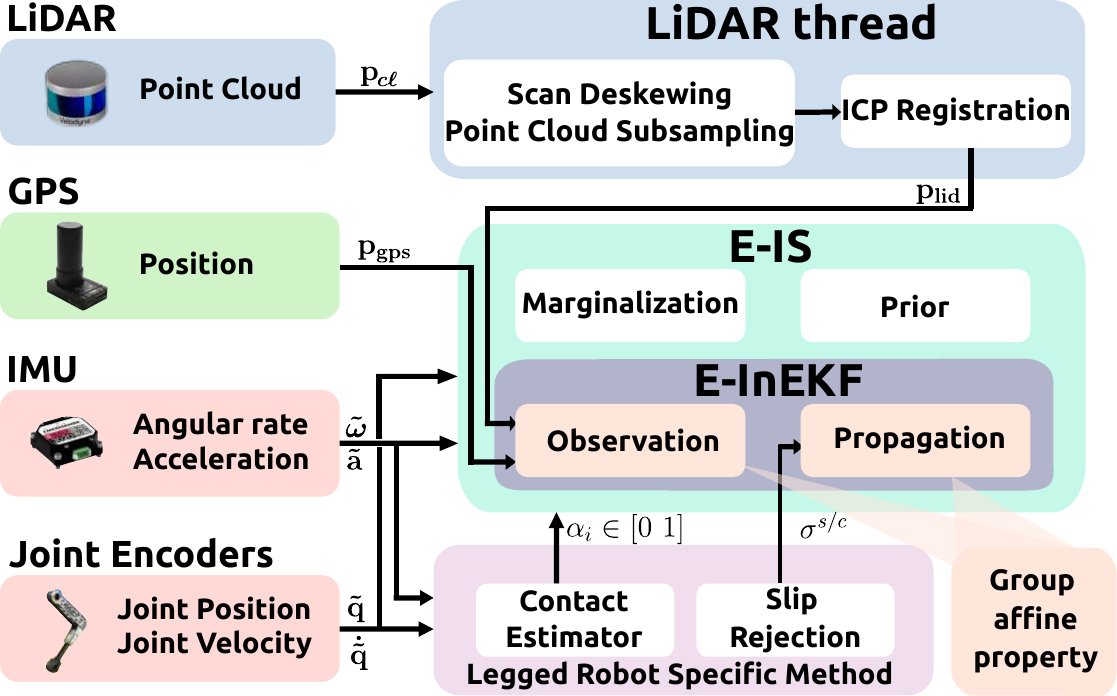}
    \caption{The structure of the two proposed frameworks. 
    The E-InEKF and E-IS have the propagation and observation modules, but the E-IS also includes the prior and marginalization modules.}
    \label{fig:is_structure}
    \vspace{-0.5cm}
\end{figure}

\subsection{E-InEKF: LiDAR-GPS fused Invariant EKF}
\label{subsec:inEKF}
Building upon the contact-aided InEKF in~\cite{hartley2020contact},
we extend the measurement model to include 
LiDAR-odometry and GPS observations.
Starting from~\eqref{eq:kine_meas},~\eqref{eq:lid_meas}, and~\eqref{eq:gps_meas}, we define 
an aggregated observation vector $\mathbf{Y} = [\mathbf{Y^\top_{kin}}, \quad \mathbf{Y^\top_{lid}}, \quad \mathbf{Y^\top_{gps}}]^\top$, 
and a corresponding constant vector $\mathbf{b} = [\mathbf{b^\top_{kin}}, \quad \mathbf{b^\top_{lid}}, \quad \mathbf{b^\top_{gps}}]^\top$. 
Rewritten in matrix form, the right-invariant observation model becomes:
\begin{equation}
\underbrace{\begin{bmatrix}
\mathbf{Y_{kin}} \\
\mathbf{Y_{lid}} \\
\mathbf{Y_{gps}} \\
\end{bmatrix}}_{\mathbf{Y}}
= \underbrace{\begin{bmatrix}
\mathbf{X^{-1}}  & \mathbf{0} & \mathbf{0} \\
\mathbf{0} & \mathbf{X^{-1}}  & \mathbf{0} \\
\mathbf{0} & \mathbf{0} & \mathbf{X^{-1}} \\
\end{bmatrix}}_{\mathbf{X_{aug}}^{-1}} \underbrace{\begin{bmatrix}
\mathbf{b_{kin}} \\
\mathbf{b_{lid}} \\
\mathbf{b_{gps}} \\
\end{bmatrix}}_{\mathbf{b}}
+ \underbrace{\begin{bmatrix}
\mathbf{V_{kin}} \\
\mathbf{V_{lid}} \\
\mathbf{V_{gps}} \\
\end{bmatrix}}_{\mathbf{V}}
\label{eq:iekf_meas}
\end{equation}
Here $\mathbf{X_{aug}}$ is the augmented state matrix, with the state $\mathbf{X}$ appearing on the diagonal.

The linear update equations follow as described in~\Cref{subsec:propagation}, with the Jacobians defined as $\mathbf{H} = [\mathbf{H_{kin}^\top} \quad \mathbf{H_{lid}^\top} \quad \mathbf{H_{gps}^\top}]^\top$, 
and the noise covariance given by the block-diagonal ($\mathbf{blkdiag}$) matrix $\mathbf{N} = \mathbf{blkdiag}(\mathbf{N_{kin}},\mathbf{N_{lid}}, \mathbf{N_{gps}})$, where the individual Jacobians are
$\mathbf{H_{kin}} = [
\mathbf{0}_{3,3} , \mathbf{0}_{3,3} , -\mathbb{I}_{3} , \mathbb{I}_{3}
],$ and
$\mathbf{H_{lid}} = \mathbf{H_{gps}} = [
\mathbf{0}_{3,3} , \mathbf{0}_{3,3} , -\mathbb{I}_{3} , \mathbf{0}_{3,3}]$, and the individual covariances are
$\mathbf{N_{kin}}=\mathbf{R}_\mathrm{t} \mathbf{J_p}(\tilde{\mathbf{q}}_\mathrm{t}) \Sigma(\mathbf{w}^{\mathbf{q}}_\mathrm{t})  \mathbf{J^\top_p}(\tilde{\mathbf{q}})_\mathrm{t} \mathbf{R}_\mathrm{t}^\top$, 
$ \mathbf{N_{lid}}=\mathbf{R}_\mathrm{t} \mathbf{J_{{lid}}} \Sigma(\mathbf{w}^{\mathbf{lid}}_\mathrm{t}) \mathbf{J_{{lid}}^\top} \mathbf{R}_\mathrm{t}^\top$, and 
$ \mathbf{N_{gps}}=\mathbf{R}_\mathrm{t} \mathbf{J_{{gps}}} \Sigma(\mathbf{w}^{\mathbf{gps}}_\mathrm{t}) \mathbf{J_{{gps}}^\top} \mathbf{R}_\mathrm{t}^\top$.
\noindent
In these expressions, $\Sigma(\mathbf{w}^{\mathbf{q}}_\mathrm{t})$, 
$\Sigma(\mathbf{w}^{\mathbf{lid}}_\mathrm{t})$, and $\Sigma(\mathbf{w}^{\mathbf{gps}}_\mathrm{t})$ are the covariance matrices of the joint position, LiDAR position, and GPS position noise, respectively.
\looseness=-1

\subsection{E-IS: LiDAR-GPS fused Invariant Smoother}
\label{subsec:is}
The E-IS is formulated such that the \textit{observation} term in the cost function~\eqref{eq:is_cost} incorporates global position data $\mathbf{p}_\mathbf{lid,t}$ and $\mathbf{p}_\mathbf{gps,t}$ from LiDAR and GPS, respectively.
These global measurements correct the state by comparing the robot's estimated position with observed reference points.

From~\eqref{eq:is_cost}, we continue with the LiDAR observation cost, by expressing the equation for the \textit{LiDAR measurement model} in a right-invariant form, similar to~\eqref{eq:lid_meas}, utilizing $\mathbf{p_{lid}}$, which represents the LiDAR position in the robot's body frame.
Based on this formulation, we compute the LiDAR residual function, which is subsequently used to update the global robot position. Specifically, the residual $\mathbf{r_{lid}}$, the Jacobian $\mathbf{J_{lid}}$, and the covariance matrix $\mathbf{\Sigma_{lid}}$ are defined as:
$
\mathbf{r_{lid}} = \mathbf{X}\mathbf{Y_{lid}} - \mathbf{b_{lid}}
$, \quad
$
\mathbf{J_{lid}} = [\mathbf{b_{lid}}^\odot \hspace{0.2cm} \mathbf{0}_{6,6}]
$, \quad and
$
\mathbf{\Sigma_{lid}} = \mathbf{X} \mathbf{\Sigma_{lid}(w^{lid}_t)} \mathbf{X}^\top
$
where the definition of $\mathbf{Y_{lid}}$ and $\mathbf{b_{lid}}$ is the same as in~\Cref{eq:lid_meas}. 

Similarly, the residual for GPS is obtained from~\eqref{eq:gps_meas}, using the same approach as for the kinematic and LiDAR observations. The GPS residual $\mathbf{r_{gps}}$, its Jacobian $\mathbf{J_{gps}}$, and the corresponding covariance $\mathbf{\Sigma_{gps}}$ are defined as
$
    \mathbf{r_{gps}} = \mathbf{X}\mathbf{Y_{gps}} - \mathbf{b_{gps}}
$, 
$
    \mathbf{J_{gps}} = [\mathbf{b_{gps}}^\odot \hspace{0.2cm} \mathbf{0}_{6,6}]
$, and
$
    \mathbf{\Sigma_{gps}} = \mathbf{X} \mathbf{\Sigma_{gps}(w^{gps}_t)} \mathbf{X}^\top
$
where $\mathbf{Y_{gps}}$ and $\mathbf{b_{gps}}$ are defined in~\Cref{eq:gps_meas}.

\Cref{fig:is_structure} provides a schematic overview of the two frameworks, illustrating their shared components and emphasizing the primary differences between them. The diagram highlights the common underlying structure while delineating where the two approaches are different.
\Cref{fig:is_structure} also shows the legged robot-specific methods from~\cite{is_legged,kim2021slip} that we incorporated to enhance the E-InEKF and E-IS. Specifically, a contact estimator based on the momentum observer of~\cite{de2005sensorless} determines the contact states $\alpha_{i}$ for the $i$-th contact, while to handle dynamic events such as slipping, we adopt a slip rejection method that adjusts the covariance of contact-related observations~\cite{kim2021slip}.

\section{Experimental Results}
\label{sec:experiments}
In this section, we present the experimental results of the E-InEKF and E-IS algorithms, using the quadruped robot KAIST Hound2. These experiments were conducted in two distinct scenarios: indoor and outdoor environments, as shown in \Cref{fig:indoor_exp} and \Cref{fig:outdoor}. All results were obtained from offline processing and analysis of the collected sensor data. Also, we underline that our state estimators are odometry systems, with no loop closures applied to reduce drift in the trajectory estimation.
\looseness=-1

For the indoor experiment, the ground truth pose was provided by a Vicon motion capture system. For the outdoor experiment, ground truth was obtained from a Holybro Real-Time Kinematic (RTK) GPS with a helical antenna.

As baselines, we adopted the proprioceptive-only invariant state estimation methods P-InEKF~\cite{hartley2020contact} and P-IS~\cite{is_legged}, as well as the state-of-the-art LiDAR-based methods LIO-SAM~\cite{shan2020lio} and FAST-LIO2~\cite{xu2022fast}. The performance of the proposed algorithms was quantified using two widely recognized metrics: the mean Absolute Trajectory Error (ATE), which assesses global pose estimation accuracy, and the mean Relative Position Error (RPE), which evaluates local consistency in pose estimation. 

\subsection{Indoor Experiments}
\label{subsec:indoor}

\begin{figure}[t]
    \centering
    \includegraphics[ width=0.48\textwidth]{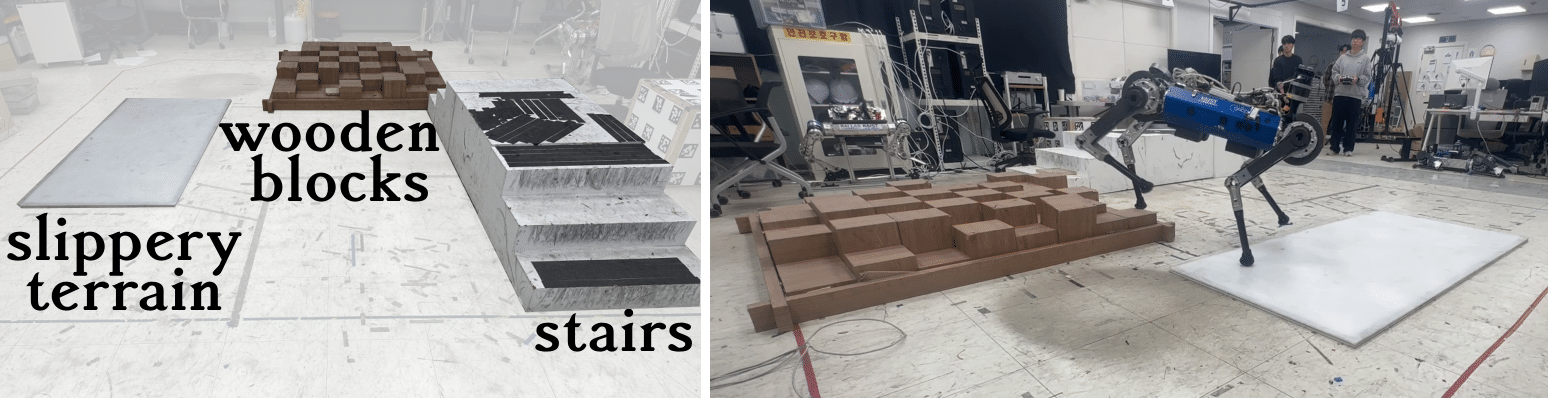}
    \caption{The indoor experiment involved large impacts and significant changes in the robot’s height, as illustrated in the figures.}
    \label{fig:indoor_exp}
    \vspace{-0.2cm}
\end{figure}

\begin{figure*}[t]
    \centering
    \includegraphics[width=0.98\textwidth]{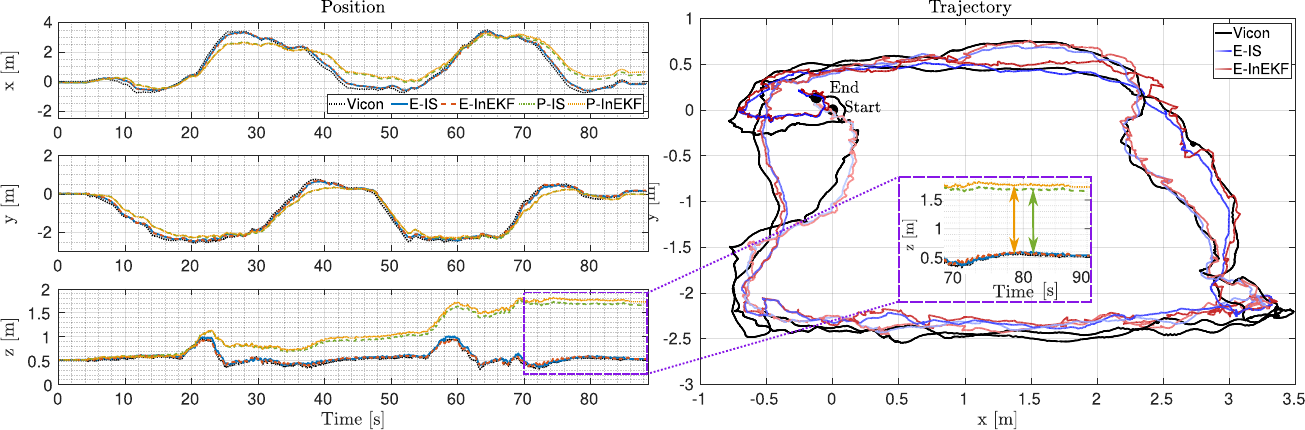}
    \caption{The results for the indoor experiment, where the robot is prone to experience high contact impacts, show that the proposed frameworks, E-IS and E-InEKF, mitigate the position drift, especially in the z-axis, compared to the proprioceptive-only methods, P-IS and P-InEKF. In the purple dotted box, the position in the z-axis of P-IS and P-InEKF is indicated by the green and yellow arrows. Vicon is used as ground truth. In the right plot, the E-IS and E-InEKF trajectories use a color gradient: the lighter sections indicate the start of the estimates, while the darker sections indicate the end.}
    \label{fig:indoor_results}
    \vspace{-0.5cm}
\end{figure*}

In the indoor experiment, the robot equipped with a Velodyne VLP16 LiDAR sensor performed walking tests in a controlled environment. Since the GPS is not informative in indoor settings, we evaluated the performance of E-IS and E-InEKF, which employ kinematics, IMU, and LiDAR data without GPS data. We refer to E-IS and E-InEKF as \textbf{E-IS (w/o GPS)} and \textbf{E-InEKF (w/o GPS)}, respectively, to indicate that they exclude the use of GPS in the indoor setting. 

The indoor environment simulates real-world challenging terrain, as shown in~\Cref{fig:indoor_exp}.
The testing area consists of wooden blocks, steps terrains, and slippery surfaces coated with boric acid on acrylic plates. While traversing these terrains, the robot experiences varying heights and high impacts during contact, which affects positional drift~\cite{ou24legkilo}.

The estimation results of the indoor experiments, as shown in~\Cref{fig:indoor_results}, clearly demonstrate the effectiveness of integrating LiDAR odometry into the invariant estimator through the proposed LiDAR observation model. This integration substantially reduced drift in the z-position, a critical improvement validated by the mean errors summarized in~\Cref{tab:indoor_comparison}. In contrast, the proprioceptive-only methods exhibit a higher susceptibility to positional drift. These findings underscore that, while the proprioceptive-only approaches reveal certain limitations, incorporating LiDAR data into the estimator significantly mitigates these issues, resulting in more robust and accurate state estimation in indoor environments.
\looseness=-1

Furthermore, both E-IS (w/o GPS) and E-InEKF (w/o GPS) recorded lower ATE and RPE than the LiDAR-based baselines. Since these methods fuse IMU and LiDAR data, without taking into account leg kinematics, their accuracy might be compromised by high-impact contacts. 
High-impact or unexpected motions, such as the slippage, caused larger errors in LIO-SAM and FAST‑LIO2, and these errors would likely increase further under even more severe impacts. In contrast, our proposed method maintained consistently robust performance by leveraging the incorporation of leg kinematics as an additional measurement.

The variations in terrain heights and high-impact contacts were critical features of this experiment, emphasizing the importance of incorporating exteroceptive sensors, such as LiDAR, to address positional drift along the z-axis. Positional drift along this axis is inherently unobservable when relying solely on proprioceptive measurements, as previously analyzed in~\cite{bloesch2013state}. 
Our proposed frameworks successfully integrate LiDAR into IS and InEKF, demonstrating how external sensors mitigate drift in invariant estimators~\cite{is_legged,hartley2020contact} and improve pose estimation in challenging conditions. 

  \begin{table}[!b]
    \begin{center}
    \caption{ATE and RPE statistics over 1 m}
    \label{tab:indoor_comparison}
    \resizebox{\columnwidth}{!}{%
    \begin{tabular}{c | c c | c c | c c }
        \toprule
        \textbf{Indoor} & E-InEKF & E-IS & LIO-SAM & FAST-LIO2 & P-InEKF & P-IS\\
         & (w/o GPS) & (w/o GPS) &  &  &  & \\
        \midrule
        \rowcolor{gray!15}
        ATE [m] & 0.24 & \textbf{0.23} & 0.32 & 0.30 & 0.90 & 0.85 \\
        RPE [m] & 0.09 & \textbf{0.08} & 0.15 & 0.16 & 0.21 & 0.20 \\
        \bottomrule
    \end{tabular}
    }
    \end{center}
    \end{table}

\subsection{Outdoor Experiments}
\label{subsec:outdoor} 
\begin{figure}[t]
    \centering
    \includegraphics[width=0.48\textwidth]{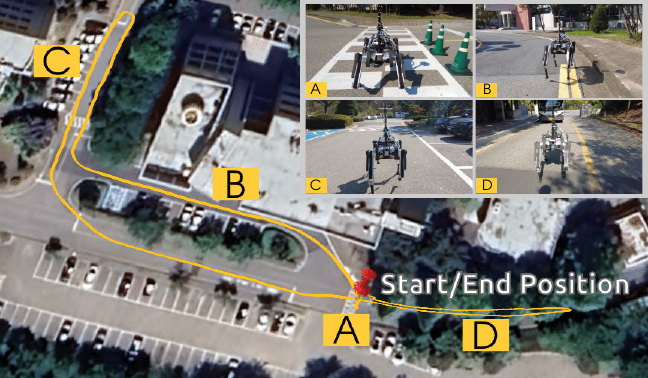}
    \caption{The illustration of the outdoor experiment: Google Earth view with screenshots of the robot walking along the outdoor path, on the top right corner.}
    \label{fig:outdoor}
    \vspace{-0.5cm}
\end{figure}

To assess the frameworks' ability to reduce long-term drift and maintain robustness in outdoor environments, 
the robot traversed a long-distance path of approximately 300~\si{\meter}, as shown in~\Cref{fig:outdoor}. The experiment was designed as a closed route, with the robot returning to its starting point. For the outdoor experiments, we used the KAIST HOUND2 robot equipped with a Velodyne VLP16 LiDAR sensor and a Holybro RTK GPS. 
\looseness=-1

\begin{figure*}[t]
    \centering
    \includegraphics[width=0.98\textwidth]{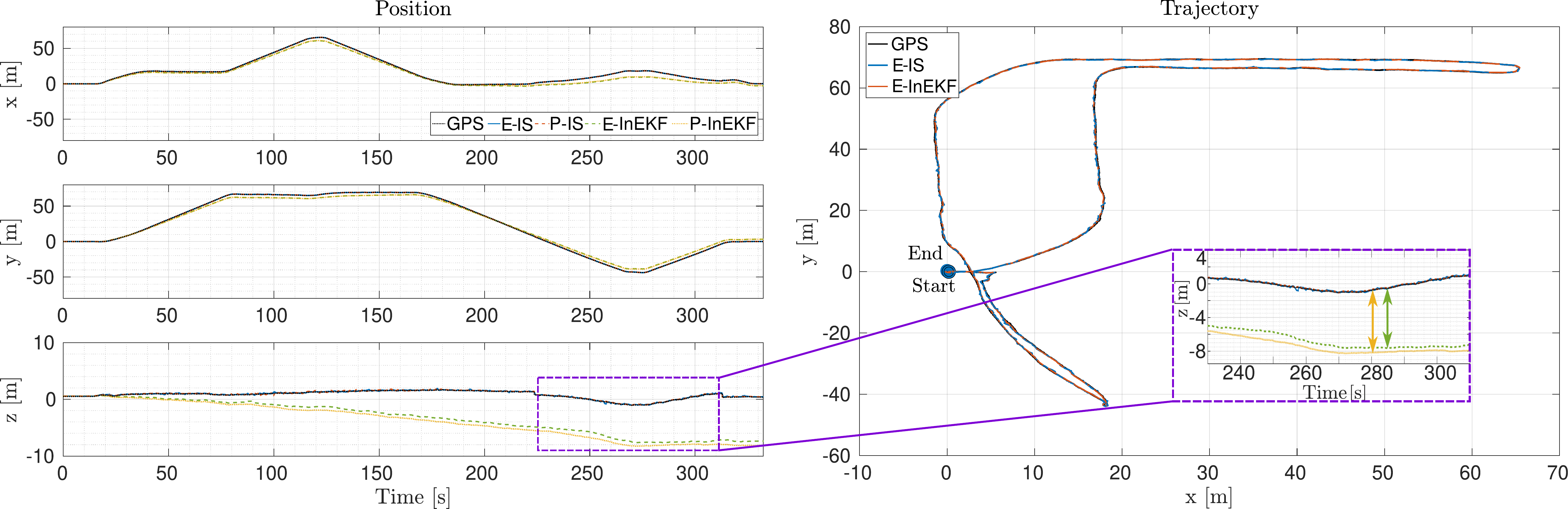}
    \caption{The results for the outdoor experiment clearly show the improvement in the z-axis drift when using LiDAR and GPS, even in long-distance operations. The drift is indicated by the green and yellow arrows, for the P-InEKF and P-IS, respectively.}
    \label{fig:outdoor_results}
\end{figure*}

As shown in~\Cref{fig:outdoor_results}, both E-IS and E-InEKF achieved lower drift compared to P-InEKF and P-IS, especially in the z-axis, thanks to the incorporation of the exteroceptive measurements.

In~\Cref{tab:outdoor_comparison}, we compared the ATE and RPE of our frameworks with those of the baselines (LIO-SAM and FAST-LIO2). We observed that the proposed frameworks, E-IS and E-InEKF, showed lower ATE and RPE compared to the baseline approaches. 
\looseness=-1

Notably, E-InEKF (w/o GPS) and E-IS (w/o GPS) yielded lower RPE than FAST-LIO2 and LIO-SAM, while FAST-LIO2’s ATE was only 2 cm lower than that of our proposed frameworks. 
Although incorporating leg kinematics can introduce potential noise that may slightly increase the drift over long trajectories, it enhances robustness, especially when LiDAR alone may be unreliable (e.g., environments with dense vegetation or high-impact scenarios), and it is still advantageous to include these measurements whenever possible.

Moreover, our group-affine observation model effectively integrates LiDAR and GPS into the invariant filter and smoother, ensuring robust long-term performance.

\begin{table}[!b]
    \begin{center}
    \caption{ATE and RPE statistics over 1 m}
    \label{tab:outdoor_comparison}
    \resizebox{\columnwidth}{!}{%
    \begin{tabular}{c | c c | c c | c c c c}
        \toprule
        \textbf{Outdoor} & E-InEKF & E-IS & LIO-SAM & FAST-LIO2 & E-InEKF & E-IS & P-InEKF & P-IS \\
         &  &  &  &  & (w/o GPS) & (w/o GPS) &  &  \\
        \midrule
        \rowcolor{gray!15}
        ATE [m] & 0.17 & \textbf{0.15} & 2.28 & 1.35 & 1.68 & 1.37 & 6.57 & 6.32 \\
        RPE [m] & 0.07 & \textbf{0.06} & 0.20 & 0.10 & 0.09 & 0.08 & 0.10 & 0.09 \\
        \bottomrule
    \end{tabular}
    }
    \end{center}
\end{table}

\section{Discussion}
\label{sec:discussion}
In this section, we analyze the results presented in~\Cref{sec:experiments}, including an ablation study and a discussion on the trade-off between E-IS and E-InEKF. The ablation study in~\Cref{sec:ablation} evaluates the contribution of each module in our approach, while the analysis in~\Cref{sec:filtvssmooth} provides valuable insights into the strengths and limitations of both methods.

\subsection{Ablation Study}
\label{sec:ablation}
To discuss the effect of exteroceptive sensors, we compare three cases within our frameworks by selectively excluding exteroceptive sensors. Specifically, we discuss E-IS, E-IS (w/o GPS), and P-IS for invariant smoothing and E-InEKF, E-InEKF (w/o GPS), and P-InEKF for invariant filtering.

As shown in~\Cref{tab:indoor_comparison} and~\Cref{tab:outdoor_comparison}, the experimental results demonstrated that the proposed methods, E-IS and E-InEKF, consistently outperformed LiDAR-based odometry. Both E-IS and E-InEKF effectively corrected positional drift along the z-axis in indoor and outdoor settings, highlighting their robustness in managing vertical displacement errors.

In the outdoor experiments, the reduced accuracy of the E-IS (w/o GPS) and E-InEKF (w/o GPS), compared to E-IS and E-InEKF, is primarily due to the use of GPS as ground truth for evaluation. 
However, even without GPS, E-IS (w/o GPS) and E-InEKF (w/o GPS) demonstrated better performance than LiDAR-based odometry in the indoor environment, where the robot could experience high-impact contacts, while they showed comparable performance to the baseline, achieving lower RPE in the outdoor experiment.
These results highlight the robustness of the proposed frameworks, even in the absence of GPS data, and underscore their ability to outperform state-of-the-art LiDAR-based odometry methods under challenging conditions.
\looseness=-1

Also, with the exteroceptive sensors of LiDAR or GPS, the positional drift of P-IS and P-InEKF is significantly reduced. Additionally, as shown in~\Cref{fig:exec_time_boxplot}, even if we incorporate the exteroceptive sensors, the computation time is not affected by the low operation rates of LiDAR or GPS, which is attributed to the loosely-coupled manner this work adopted. Notably, since the LiDAR odometry is calculated on a separate thread, we then apply the proposed observation model for LiDAR odometry into E-IS and E-InEKF, while satisfying the group-affine property.
\looseness=-1

\subsection{Comparison between filtering and smoothing}
\label{sec:filtvssmooth}
\begin{figure}[t]
    \centering
    \includegraphics[width=0.48\textwidth]{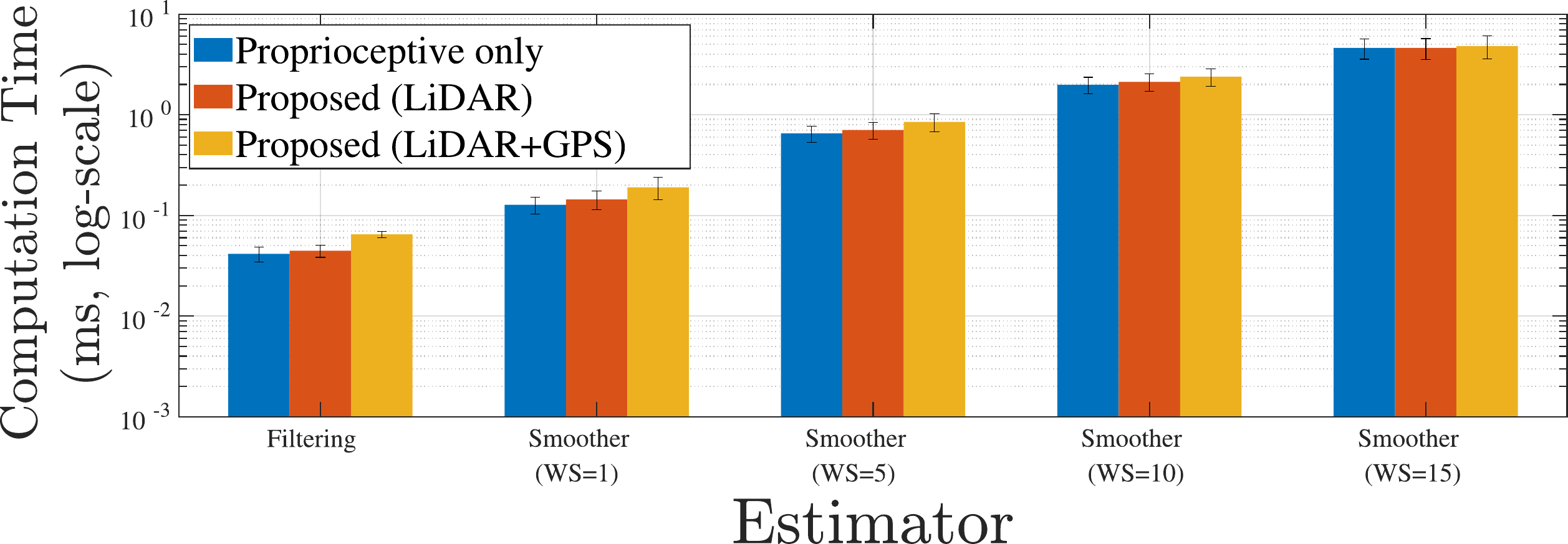}
    \caption{Comparison of average computation times for E-IS, E-IS (w/o GPS), P-IS across different window sizes (WS) and their InEKF-based counterparts (E-InEKF, E-InEKF (w/o GPS), P-InEKF). The results show that while a larger window size increases the computation times of E-IS and E-IS (w/o GPS), they remain comparable to the proprioceptive-only cases due to the loosely coupled method used in this work.}
    \label{fig:exec_time_boxplot}
    \vspace{-0.4cm}
\end{figure}

In all tests, the E-IS algorithm achieved better positional accuracy than E-InEKF by incorporating a smoothing window that exploits past states. However, E-InEKF’s reduced computational cost makes it better suited for real-time applications. As shown in~\Cref{tab:atews_comparison} and~\Cref{fig:exec_time_boxplot}, we evaluated both algorithms under different window sizes (WS = 1, 5, 10, and 15) to assess the impact of the window length on accuracy and execution time. \Cref{tab:atews_comparison} shows how the ATE decreases with larger window sizes, while~\Cref{fig:exec_time_boxplot} illustrates the corresponding increase in computation time. In all experiments, E-IS achieved higher positional accuracy than E-InEKF by incorporating state history through smoothing, while E-InEKF showed instead to be more efficient in terms of computation time.
\looseness=-1

Increasing the window size of E-IS improves accuracy but also increases the computation time. For instance, with WS = 15, E-IS achieves the lowest ATE, but has an average computation time of 4.5~\si{\milli\second} per iteration, whereas E-InEKF requires 0.06~\si{\milli\second} per iteration. Reducing E-IS's window to WS = 1 lowers the execution time to 0.18~\si{\milli\second}, with a marginal decrease in accuracy compared to E-InEKF. 

Hence, the choice between the E-IS and E-InEKF depends on the application's requirements. If real-time performance is critical, E-InEKF is preferred. Conversely, tasks demanding higher pose accuracy, such as detailed mapping or high-fidelity localization, benefit from a larger smoothing window in E-IS despite its higher computational cost.

  \begin{table}[!b]
    \begin{center}
    \caption{ATE of the proposed E-InEKF and E-IS with different WS}
    \label{tab:atews_comparison}
    \resizebox{\columnwidth}{!}{%
    \begin{tabular}{c| c c c }
        \toprule
         & Proprioceptive only & Proposed (LiDAR) & Proposed (LiDAR + GPS) \\
         \midrule
         \rowcolor{gray!15}
         Filter & 6.340664 m & 1.683859 m & 0.171617 m \\
         Smoother (WS:1) & 6.331560 m & 1.833524 m &  0.156991 m\\
         \rowcolor{gray!15}
         Smoother (WS:5) & 6.322465 m & 1.638729 m & 0.150772 m\\
         Smoother (WS:10) & 6.316255 m & 1.484747 m &  0.150680 m\\
         \rowcolor{gray!15}
         Smoother (WS:15) & \textbf{6.313716 m} & \textbf{1.376002 m} & \textbf{0.150467 m} \\
        
        \bottomrule
    \end{tabular}
    }
    \end{center}
    \end{table}

\section{Conclusion}
\label{sec:conclusion}
In this work, we presented two state estimation frameworks for quadruped robots, E-IS and E-InEKF, that combine LiDAR and GPS with the proprioceptive-based InEKF~\cite{hartley2020contact} and IS~\cite{is_legged}.
To incorporate the LiDAR and GPS into the invariant state estimators, we proposed an observation model for LiDAR and GPS that satisfies the group-affine property for the invariant estimators. Moreover, to handle the low frequency of LiDAR measurements, we employ a parallel thread to obtain the LiDAR odometry.
To the best of our knowledge, this is the first approach integrating LiDAR and GPS into invariant estimators through a group-affine observation model and a parallel-thread LiDAR odometry module. Indoor and outdoor experiments on Hound2~\cite{hound22shin} showed that our methods significantly reduce z-position drift and achieve lower RPE compared to existing proprioceptive and LiDAR-based methods~\cite{hartley2020contact,is_legged,shan2020lio,xu2022fast}.

Future work will focus on integrating RGB-D or thermal cameras and comparing them with tightly-coupled methods to better understand the trade-offs in sensor fusion.
\vspace{-0.4cm}
 
\bibliographystyle{IEEEtran}
\bibliography{bibliography.bib}

\end{document}